\documentclass{article}

\usepackage{microtype}
\usepackage{graphicx}
\usepackage{subfigure}
\usepackage{booktabs} 

\usepackage{hyperref}

\usepackage{multirow}
\usepackage{amsmath}
\DeclareMathOperator*{\argmax}{arg\,max}

\usepackage[accepted]{icml2024}

\usepackage{amsmath}
\usepackage{amssymb}
\usepackage{mathtools}
\usepackage{amsthm}

\usepackage[capitalize,noabbrev]{cleveref}

\theoremstyle{plain}

\theoremstyle{definition}

\theoremstyle{remark}

\usepackage[textsize=tiny]{todonotes}

\icmltitlerunning{Reinforcement Learning for Energy Systems Co-Optimisation}

\begin{document}

\twocolumn[
\icmltitle{Reinforcement Learning for Efficient Design and Control\\ Co-optimisation of Energy Systems}




\begin{icmlauthorlist}
\icmlauthor{Marine Cauz}{yyy1,comp}
\icmlauthor{Adrien Bolland}{yyy2}
\icmlauthor{Christophe Ballif}{yyy1}
\icmlauthor{Nicolas Wyrsch}{yyy1}
\end{icmlauthorlist}

\icmlaffiliation{yyy1}{Photovoltaics and Thin-Film Electronics Laboratory (PV-lab), Institute of Electrical and Micro Engineering (IEM), École Polytechnique Fédérale de Lausanne (EPFL), Neuchâtel, Switzerland}
\icmlaffiliation{yyy2}{Department of Electrical Engineering and Computer Science, University of Liège, Liège, Belgium}
\icmlaffiliation{comp}{Planair SA, Yverdon-les-Bains, Switzerland}

\icmlcorrespondingauthor{Marine Cauz}{marine.cauz@epfl.ch}

\icmlkeywords{Reinforcement Learning, Energy, Renewable, Optimisation}

\vskip 0.3in
]



\printAffiliationsAndNotice{}  

\begin{abstract}
The ongoing energy transition drives the development of decentralised renewable energy sources, which are heterogeneous and weather-dependent, complicating their integration into energy systems. This study tackles this issue by introducing a novel reinforcement learning (RL) framework tailored for the co-optimisation of design and control in energy systems. Traditionally, the integration of renewable sources in the energy sector has relied on complex mathematical modelling and sequential processes. By leveraging RL's model-free capabilities, the framework eliminates the need for explicit system modelling. By optimising both control and design policies jointly, the framework enhances the integration of renewable sources and improves system efficiency. This contribution paves the way for advanced RL applications in energy management, leading to more efficient and effective use of renewable energy sources.
\end{abstract}

\section{Introduction}
\label{introduction}
    \subsection{Background and motivation}
    \label{background}
        Energy systems are undergoing significant transformations to meet increasing demands for sustainability and energy efficiency, particularly through the integration of decentralised and intermittent renewable energy sources. Traditionally, these systems are developed in two distinct phases: design, which determines the optimal size of components, and control, which focuses on their optimal operation. This sequential approach, as highlighted by \cite{dranka_review_2021}, often leads to inefficiencies and missed opportunities for optimal performance. To address the increasing complexity driven by renewable integration, co-optimisation has emerged as a key approach, jointly handling design and control to enhance system reliability and affordability. Recent literature underscores the importance of co-optimising design and operation using techniques such as linear programming \cite{krishnan_co-optimization_2015, daadaa_optimization_2021, jayadev_us_2020}, stochastic models \cite{clack_linear_2015,qiu_stochastic_2017}, robust optimisation \cite{popovici_framework_2015,khojasteh_robust_2020}, and evolutionary algorithms \cite{li_optimal_2018, gjorgiev_electrical_2018, bao_optimal_2019}. Among these methods, Mixed-Integer Linear Programming (MILP) is the most commonly used but requires mathematical modelling of the system and its interactions. These methods aim to optimise performance comprehensively while addressing uncertainties and multi-objective challenges. Overall, these diverse approaches highlight both the technical challenges and the critical importance of co-optimisation in enhancing the efficiency and sustainability of energy systems \cite{sachio_integrating_2022, fazlollahi_multi-objective_2013, dranka_review_2021}.

        Data-driven methods, such as reinforcement learning (RL), have shown significant potential in computing control policies across various applications, including energy, offering a promising alternative to traditional approaches \cite{francois-lavet_introduction_2018, quest_3d_2022, perera_introducing_2020}. However, standard RL methods typically focus solely on operational control without integrating system design, limiting insights into how design changes influence outcomes. Despite its potential, RL is not fully exploited in the energy field \cite{perera_applications_2021}. Recent advancements in RL, particularly gradient-based optimisation techniques like actor-critic methods, facilitate learning control policies for complex problems, opening new opportunities.

        Building on these advancements, researchers have proposed algorithms to efficiently tackle joint design and control challenges. In \cite{schaff_jointly_2019}, the authors introduced an RL framework that optimises both design and control by maintaining a distribution over designs, using the Proximal Policy Optimization (PPO) algorithm \cite{schulman_proximal_2017} for policy training and the \textsc{reinforce} update rule for design adjustments \cite{williams_simple_1992}. This approach has been successfully applied in various robotic environments, outperforming other techniques \cite{bhatia_evolution_2022, ha_reinforcement_2019}. Alternatively, \cite{luck_data-efficient_2020} enhances adaptability for joint design and control using Soft Actor-Critic (SAC) \cite{haarnoja_soft_2018}, despite involving complex optimisation problems. The algorithm from \cite{bolland_jointly_2022} refines this approach by combining policy gradients with model-based optimisation, It was applied to systems with photovoltaic (PV) panels and battery \cite{cauz_reinforcement_2023}, though it faces limitations due to finite time horizons and on-policy nature. Other approaches \cite{chen_hardware_2020, jackson_orchid_2021} focus on learning system parameters directly, assuming the system dynamics are parameterised, but are restrictive when modelling complex energy systems where design decisions are directly related to explicit costs or rewards.

    \subsection{Contribution}
    \label{contribution}       
        Capitalising on these recent developments in policy gradient techniques, this study advances an integrated RL framework specifically tailored to address the co-optimisation challenges within energy systems. As introduced by \cite{schaff_jointly_2019}, the proposed framework employs a parametric design distribution, whose parametric nature is effective for modelling distributions over continuous supports and allows for using gradient based methods easily. This approach contrasts with most of the previous methods that employ a deterministic representation of the design variable \cite{chen_hardware_2020, jackson_orchid_2021, bolland_jointly_2022}, which can make model-free optimisation and efficient exploration challenging. Additionally, this framework distinguishes from \cite{schaff_jointly_2019} by incorporating entropy regularisation, as in \cite{haarnoja_soft_2018}, into the optimisation process to prevent convergence to local optima. Furthermore, this framework relies on a deterministic policy parameterisation, which is optimised using an off-policy actor-critic algorithm, namely Deep Deterministic Policy Gradient (DDPG) \cite{lillicrap_continuous_2019}. This allows for accommodating infinite time horizons, addressing a significant gap in methodologies \cite{bolland_jointly_2022, cauz_reinforcement_2023}. Unlike most existing studies, including \cite{schaff_jointly_2019}, the control policy training is off-policy, thereby enhancing sample efficiency by learning from a diverse range of past experiences stored in a replay buffer. Finally, this framework is also model-free, eliminating the need for a predefined mathematical model of the system, which simplifies implementation and broadens its applicability. None of the previously cited methods combine all these features.
        
        By integrating these capabilities, this approach maximises the potential of RL to address the co-optimisation of design and operation within energy systems, a challenge often overlooked in RL research. This integrated framework bridges the gap between theoretical RL research and its practical application in energy systems, establishing a new benchmark for employing RL to tackle co-optimisation challenges in the energy sector.
        
        The paper is structured as follows: Section \ref{ch7:method} details the proposed RL method, covering both control and design aspects. Section \ref{ch7:experiments} describes the energy system and experimental setup. Section \ref{ch7:results} presents the findings, with Section \ref{ch7:discussion} discussing their implications and potential impact. Finally, Section \ref{ch7:conclusion} summarizes the key insights and contributions of the research.

\section{Method}
\label{ch7:method}

    This section outlines the conventional RL approach for system control and then details the adaptations made to enable learning system designs.
    
    \subsection{Control Policy}
    \label{sec:policy}
        Formally, RL is conceptualised as an interplay between an agent and an environment. This environment is mathematically formalised as a Markov Decision Process (MDP) \cite{bellman_markovian_1957}, which is defined by its model \( \mathcal{M} \) = \( (\mathcal{S}, \mathcal{A}, T, R, p_0, \gamma) \), where \( \mathcal{S} \) denotes the state space, \( \mathcal{A} \) denotes the action space, \( T : \mathcal{S} \times \mathcal{A} \times \mathcal{S} \rightarrow [0, 1] \) denotes the transition function (i.e., \( T(s_{t+1}|s_t,a_t) \) denotes the probability of reaching a state \( s_{t+1} \) when taking an action \( a_t \) from state \( s_t \)), \( R : \mathcal{S} \times \mathcal{A} \rightarrow \mathbb{R} \) denotes the reward function (i.e., \( R(s_t,a_t) \) is the immediate reward received by taking action \( a_t \) from state \( s_t \)), \( p_0 : \mathcal{S} \rightarrow [0, 1] \) denotes the initial distribution, \( \gamma \in [0, 1) \) denotes the discount factor (i.e., \( \gamma \) models the importance of future rewards, with a lower value placing more emphasis on immediate rewards). Within the MDP framework, the agent's objective is to find a policy, \( \pi \in \Pi \), namely a conditional distribution over actions that can be used to take actions in each state by sampling. The optimal policy denoted \( \pi^* \)  maximises the cumulative reward, called expected return of the policy:  \( \mathbb{E} \left[ \mathcal{R}_t \right] \), such as  \( \mathcal{R}_t = \sum_{t=0}^{\infty} \gamma^t R(s_t, a_t) \).

        Actor-critic algorithms combine policy gradient and value-based methods for efficient policy learning and evaluation \cite{francois-lavet_introduction_2018}. The actor proposes actions based on a policy $\pi$ modelled by a neural network with parameters $\theta$, while the critic evaluates these actions by estimating value functions. This mechanism allows for ongoing refinement of the policy based on the critic's feedback and updating the critic as the policy changes. Among the various actor-critic implementations, DDPG \cite{silver_deterministic_2014, lillicrap_continuous_2019} stands out due to its off-policy nature, meaning the policy can be improved using trajectories where actions are taken from another policy, and is suitable for environments with continuous action spaces. The critic approximates the state-action value function \( Q^{\theta}(s,a) \), aiding in policy update gradients. To ensure stable learning, DDPG employs target networks for temporal-difference learning benchmarks and adds Gaussian noise to policy outputs for sufficient exploration.   

    \subsection{Design Policy}
        Conventional RL typically focuses on optimising a control policy \( \pi^*_{\theta} \) for a fixed system design. Building on this primary objective, this study explores both the design space \( X \) and control strategies to identify an optimal system design \( x^* \) and its corresponding control policy \( \pi^*_\theta(a_t|s_t,x^*) \). To each design \( x \in X\) corresponds a different MDP, as defined in Subsection \ref{sec:policy}. The objective is to maximise the expected return over a design distribution, effectively co-optimising design and control to enhance overall system performance. The proposed RL framework extends the traditional control policy optimisation by incorporating a probability distribution \( p_{\phi}(x) \) over potential designs \( x \in X \). The learnable parameters $\phi$ represent the parameters of this design distribution. The ultimate goal is to find the optimal parameters \( \phi^* \) and \( \theta^* \) that jointly maximise the expected discounted reward:
        \begin{align}
            \phi^*, \theta^* = \argmax_{\phi, \theta} \mathop{\mathbb{E}}_{x \sim p_\phi(\cdot)} \left[ \mathop{\mathbb{E}}_{\substack{s_0 \sim p_0(\cdot) \\ a_t \sim \pi_\theta (\cdot | s_t,x) \\ s_{t+1} \sim T(\cdot | s_t, a_t)}} \left[ \mathcal{R}_t \right] \right]
        \end{align}

        The co-optimisation framework is designed to maximise the expected discounted reward by effectively integrating system design and control. It is compatible with any standard RL algorithm, however, this implementation specifically uses the DDPG algorithm. This algorithm adapts the control policy \(\pi_\theta \) to maximise expected returns across a range of designs drawn from the design probability distribution \(p_\phi\). Each training iteration consists of two concurrent processes:
        \begin{itemize}
            \item The control policy \(\pi_\theta \) is refined using gradient ascent to enhance reward expectations over the sampled designs.
            \item Simultaneously, the design distribution \(p_\phi\) is updated to increase the likelihood of designs that yield higher performance under the current policy.
        \end{itemize}
       
        \begin{algorithm}[tb]
           \caption{Co-optimisation of design and control}
           \label{algo:framework}
        \begin{algorithmic}
           \STATE Initialise actor \(\pi_{\theta}(s, x)\) and critic \(Q_{\theta}(s, a, x)\)
            \STATE Initialise target networks and replay buffer (capacity \( N \))
            \STATE Initialise design distribution \( p_\phi(x) \)
            \REPEAT
                \STATE Sample designs $\{x_1, \ldots, x_d\}$ from $p_\phi(x)$
                \STATE Compute expected return $R_i$ for each $x_i$ with DDPG
                \STATE Update critic by minimising the loss:
                \STATE \(L(\theta^Q) = \frac{1}{N} \sum_{n=0}^{N-1} (y_n - Q_{\theta}(s_n, a_n, x_n))^2\)
                \STATE Update actor by one step of gradient descent:
                \STATE \(\quad \nabla_{\theta^\pi} J(\theta^Q, \theta^\pi) \approx \)
                \STATE \(\frac{1}{N} \sum_{n=0}^{N-1} \nabla_{\theta^\pi} Q_{\theta}(s_n, \pi_{\theta}(s_n, x_n), x_n) \)
                \STATE Update target networks
                \STATE Compute the loss function for the design update:
                \STATE \( \mathcal{L}(\phi) = -\frac{1}{d} \sum^{d}_{i=1} \left( \log p_{\phi}(x_i) \cdot R_{i, t} - \lambda  \cdot \log p_{\phi}(x_i) \right) \)
                \STATE Update $p_\phi$ by minimising the loss with respect to $\phi$
            \UNTIL{End of training}
        \end{algorithmic}
        \end{algorithm}
        
        Algorithm \ref{algo:framework} describes the co-optimisation procedure. It starts with the initialisation of the design distribution, fostering a wide-ranging exploration of designs. During training, the framework adjusts the policy parameters $\theta$ and the design parameters $\phi$ to gradually phase out less effective designs, allowing the policy to specialise and focus on a narrowing set of promising designs. As a result, the variance within the design distribution \( p_\phi\) decreases, guiding the system towards the convergence on an optimal design $x^*$ and associated policy \( \pi^*_\theta\), thereby maximising the overall system performance.

        In comparison with the framework proposed by \cite{schaff_jointly_2019}, two notable modifications in the design distribution enhance its suitability for energy systems. Firstly, instead of using a Gaussian Mixture Model to parameterise the design distribution \( p_\phi\), which may require clipping to ensure physical feasibility, this framework employs a log-normal mixture model. This model inherently restricts the design space to \(X = \mathbb{R}^+ \), ensuring all design values remain within physically feasible limits for energy systems. The mixture model parameters, including the mean and variance of each log-normal component and their respective (unscaled) weights, are updated using stochastic gradient ascent based on the \textsc{reinforce} gradient estimates \cite{williams_simple_1992}. The second modification introduces entropy regularisation to the design distribution to mitigate the risk of local optima, a common challenge in energy system optimisations as noted in \cite{cauz_reinforcement_2023}. Initially, the design distribution is set with random means and high variance to encourage diverse explorations. Additionally, an entropy term is added in the loss function \cite{ahmed_understanding_2019}, which gradually decreases to strategically reduce exploration over time. There is no straightforward computation of the entropy of a log-normal mixture model, hence the entropy is estimated by extending the return with the log probability of the design samples, bypassing the need for computationally intensive methods.

\section{Experiments}
\label{ch7:experiments}

    This section describes the experimental set up to evaluate the proposed framework on a building-scale PV-battery system. The aim is to minimise total electricity costs by optimising both the investment in system components, namely the design parameters, and the operational strategies for storage management, meaning the control policy. Operational costs are derived from grid interactions required to meet building energy demands. Performance is measured against the average expected return of the system's total cost, reflecting the economic impact of chosen design and control strategies. For comparative analysis, the RL co-optimisation is benchmarked against traditional approaches highlighted in Section \ref{introduction}. First, it is compared to a MILP approach for selecting the best design, followed by an RL technique for determining the optimal policy for this design. Second, it is compared to expert rule-based controllers.

    \subsection{Building-Scale System}
        The system is a building-scale energy system within an office setting, equipped with a PV installation and a stationary lithium-ion battery to satisfy its electricity requirements. The system also features a bidirectional EV (Electric Vehicle) charging point, whose usage is stochastically modelled based on typical patterns. Moreover, the building is connected to the electrical grid, subject to dynamically varying electricity prices. The main objective is to determine the optimal design for the PV installation and the battery capacity, while simultaneously developing an optimal control policy for battery and EV management. This aims to minimise the total cost of ownership, encompassing both capital and operational expenses, as well as grid costs. The design and control model of this system is formulated using an MDP, which is described in detailed in Appendix \ref{ap:A}.

        The model is trained on a historical one-year dataset of normalised PV production and electrical consumption, divided into training and validation sets to capture seasonal fluctuations. This dataset is supplemented with synthetic data for dynamic grid tariffs and EV arrival times. Details of both datasets are provided in Appendix \ref{ap:A}. The MDP's time horizon is truncated after $T=168$ hours (one week), with long-term dependencies captured via bootstrapping in the critic training. Ideally, the time horizon would cover an entire year or the system's lifecycle to capture seasonal production and consumption variations and potential equipment degradation. Performance is regularly evaluated in two ways, (i) across the full training dataset, corresponding to $T=8088$ hours, to assess long-term effectiveness, and (ii) across the full validation dataset, corresponding to $T=672$ hours, to avoid overfitting.

     \subsubsection{Experiment setup}
        The actor \( \pi_\theta\) and critic \( Q_\theta\) are both implemented using neural networks, each consisting of two hidden layers with 256 neurons and ReLU activation functions. For the actor network, a tanh activation function is applied in the final layer to map the output to the action space. The critic network concatenates the state and action at the input layer, with a linear activation function in the output layer. To facilitate integration with existing RL libraries, the design parameters \( x \) are appended to the state variables before they are inputted into the control network.

        The design distribution, \( p_\phi (x) \), is modelled as a log-normal mixture with three components, each parameterised with two design parameters. The means, variances and weights of each component are initialised randomly within the interval $\left[0,1\right[$, set high and uniformly distributed, respectively, to ensure the distribution covers a large range of \( \mathbb{R}^+ \). The entropy weight linearly decreases throughout training and reaches zero during the last half of iterations.
        
        The framework employs the DDPG algorithm to train the control policy \( \pi_\theta \). Each iteration consists of a batch of 32 episodes, each lasting $T=168$ hours, i.e., one week. Additionally, at every iterations, a set of design parameters is sampled and evaluated with the current policy across the full training dataset to monitor performance over a duration close to one year, i.e., $T=8088$ hours. To prevent gradient explosion during training, gradient clipping is implemented. Moreover, the performance are evaluated every iterations still with a batch of 32 episodes across the full validation dataset, i.e., $T=672$, with the current design distribution and control policy. Finally, the medians and quartiles of the design parameter distribution are computed at the end of training, after 500 iterations. To ensure reliability and account for variability in initialisation, all experiments are conducted using 30 different seed values ranging from 0 to 30.

    \subsubsection{Rule-based baseline}
        A rule-based baseline is established as a fixed control policy, focusing solely on optimising the design. This setup allows for a direct comparison between joint optimisation using DDPG and simple design optimisation under a given expert policy. The rule-based discharges the stationary battery when consumption exceeds PV production and charges it when production is higher than consumption. The bidirectional EV's battery, when available, follows the same logic to augment the system’s capacity. This rule-based controller operates within the same MDP environment but does not require a training phase, as it involves no trainable control parameters \( \theta \). Performance evaluations of the system’s design under this controller are conducted over 500 iterations, focusing exclusively on updating the design parameters \( \phi \), given that the control policy is static and predetermined. This experiment is referred to as the \textbf{design-only scenario}, as only the design parameters are trained, without assuming perfect foresight.

    \subsubsection{MILP baseline}
        MILP is the most widely used tools for designing energy systems  \cite{dranka_review_2021, perera_applications_2021}. This approach requires mathematical modelling of the system and its interactions, assuming a perfect foresight approach. In this study, an environment formulated as a mathematical program consisting of constraints and objectives is developed, similar to the MDP presented in Appendix \ref{ap:A}. This formulation allows for computing the optimal design, which is then controlled by a policy learned using DDPG for this particular design. This methodology enables benchmarking the proposed co-optimisation framework against a two-step baseline where the design is initially computed using MILP over the full training dataset, and subsequently controlled with DDPG. This experiment is referred to as the \textbf{best two-step scenario} because it involves a co-optimisation using the best response two-step algorithm.
      
        Moreover, a final scenario, referred to as the \textbf{fixed scenario}, is computed based on the rule-based control. The rule-based policy is implemented as constraints in the MILP (actions are constrained based on the state), allowing for the computation of the optimal design for this fixed control policy using MILP. This experiment provides a static performance, assuming fixed design and control. Since there are no training parameters, this scenario is computed to verify that the solutions provided exceed this fixed baseline.

\section{Results}
\label{ch7:results}
    This section details the performance of the proposed framework in co-optimising the design and operation of a building-scale PV-battery system.
       
    \subsection{Training Dynamics}
        \begin{figure}[ht]
            \vskip 0.2in
            \begin{center}
            \centerline{\includegraphics[width=\columnwidth]{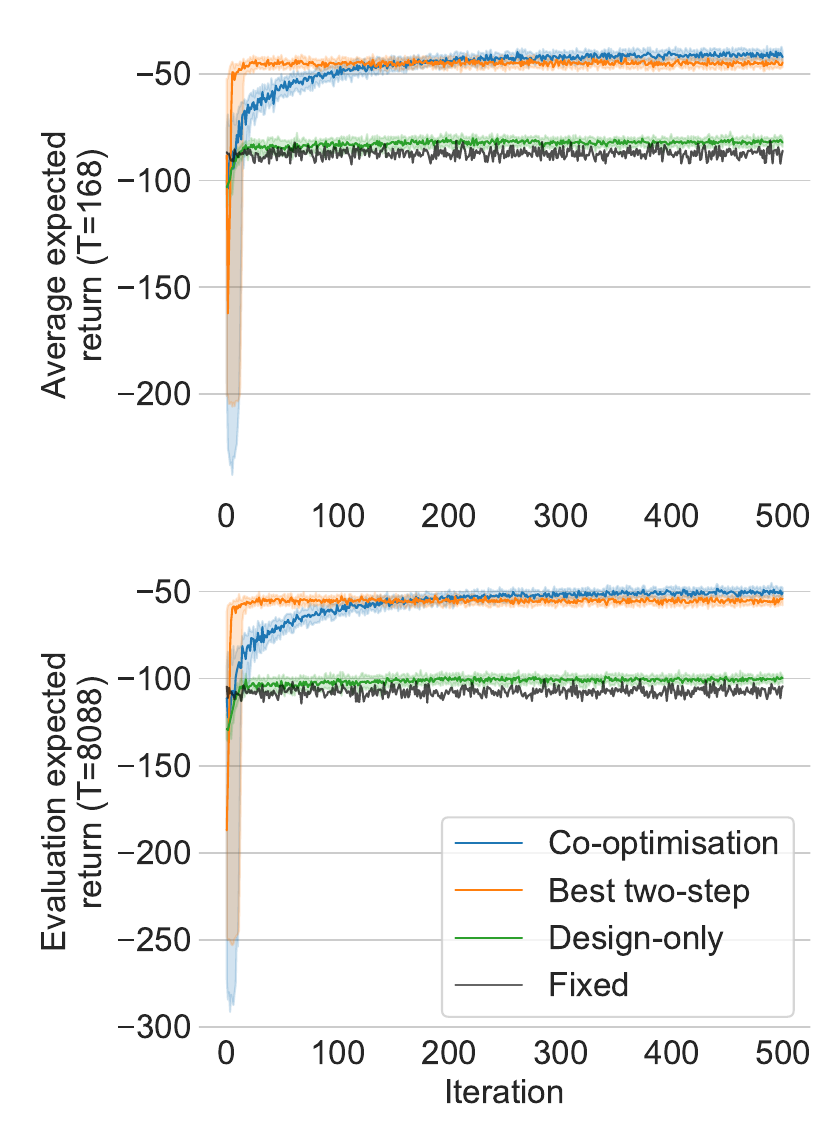}}
            \caption{Training performances, over 500 iterations, for the co-optimisation (blue), best two-step (orange), and design-only (green) scenarios. Experiments were conducted using seed values ranging from 0 to 30, with the figure showing the median and quartiles. The top subplot illustrates the evolution of average expected returns on $T$=168, i.e., the effective training. The bottom subplot assesses the average expected return throughout the full training dataset on $T$=8088, i.e., the long-term performance.}
            \label{fig:control_training}
            \end{center}
            \vskip -0.2in
        \end{figure}

    Figure \ref{fig:control_training} tracks the performances over 500 iterations during training of (i) the co-optimisation using DDPG (blue), (ii) the best two-step optimisation using DDPG for control with a fixed design derived from the MILP (orange), (iii) the design-only scenario using a rule-based control policy while optimising the design distribution (green), and (iv) the fixed scenario corresponding to the solution provided first by the MILP design computed with the rule-based constraints and then by applying the rule-based control policy (black). For all scenarios, 30 experiments are conducted with different seeds ranging from 0 to 30. Figure \ref{fig:control_training} reports the median and quartiles of the return during each learning procedure. In the best two-step scenario, the design parameters resulting from the MILP are fixed at $6$ kWp for PV and $14$ kWh for battery capacity. In the fixed scenario, the design parameters resulting from the MILP are fixed at $3$ kWp for PV and $14$ kWh for battery capacity, indicating that the integration of the rule-based constraint within the MILP constraints reduces the optimal PV power by half.  
    
    The top subplot of Figure \ref{fig:control_training} illustrates the weekly average expected returns, computed from designs sampled from the current design distribution in the co-optimisation and design-only scenarios, across batches of 32 episodes, each lasting $T=168$ hours. For these two scenarios, design parameters $\phi$ are updated during training and then the weekly average expected return stabilises by 500 iterations at mean values of $-41.7$ and $-82.9$ at the last iteration, respectively, as reported in Table \ref{tab:perf}. In the best two-step scenario, due to its static design, training converges faster, with results stabilising around $-44.8$. In the fixed scenario, since the design was previously computed using MILP and the control policy is predefined, there is no further optimisation, and it converges to $-85.9$. The variations are linked to the samples in the initial states. The bottom subplot of Figure \ref{fig:control_training} evaluates long-term performance over a batch of 32 episodes, each with a duration of the entire training dataset, $T=8088$ hours. The difference in performance between the co-optimisation and the best two-step scenario remains similar, with results converging to $-49.8$ and $-54.8$, respectively. In the design-only and fixed scenarios, the results slightly decrease compared to the weekly results, converging to $-101.1$ and $-104.3$, respectively. This assessment confirms that the co-optimisation maintains performance over extended operational periods, which is essential for the infinite horizon characteristic of energy systems.


    \begin{table}[ht]
        \caption{Average expected returns and standard deviation at the last iteration over the 30 seed experiments for training, long-term performance, and validation in the co-optimisation, best two-step, design-only, and fixed scenarios.}
        \label{tab:perf}
        \vskip 0.15in
        \begin{center}
        \begin{small}
        \begin{sc}
        \begin{tabular}{lccc}
        \toprule
        Scenario & Training & Long-term & Validation \\
                & $T$=168 & $T$=8088 & $T$=672 \\
        \midrule
        \textbf{Co-optim.}   & \textbf{-41.7$\pm$3.2}  & \textbf{-49.8$\pm$4.9}  & \textbf{-50.1$\pm$2.1}  \\
        
        Best 2-step & -44.8$\pm$4.4  & -54.8$\pm$4.1  & -54.5$\pm$0.0  \\
        Design-only &  -82.9$\pm$6.5 & -101.1$\pm$7.9 & -97.1$\pm$7.4  \\
        Fixed       & -85.9$\pm$0.0  & -104.3$\pm$0.0 &  -99.6$\pm$0.0 \\
        \bottomrule
        \end{tabular}
        \end{sc}
        \end{small}
        \end{center}
        \vskip -0.1in
    \end{table}

    \subsection{Evaluation Process}   

        \begin{figure}[ht]
            \vskip 0.2in
            \begin{center}
            \centerline{\includegraphics[width=\columnwidth]{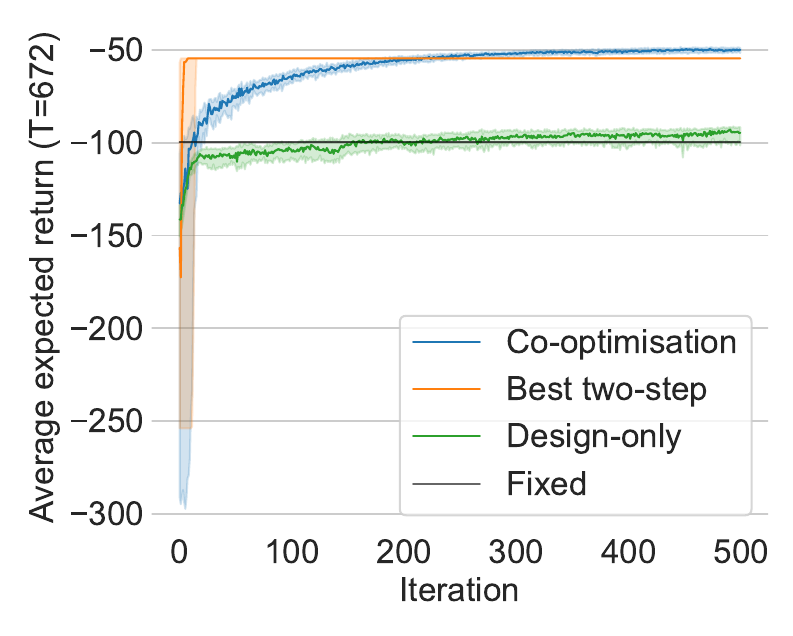}}
            \caption{Validation performances, over 500 iterations, for the co-optimisation (blue), best two-step (orange), design-only (green), and fixed (black) scenarios. Experiments were conducted using seed values ranging from 0 to 30. The figure shows the median and quartiles of the average expected return computed over the entire validation dataset, i.e., $T$=672.}
            \label{fig:control_validation}
            \end{center}
            \vskip -0.2in
        \end{figure}
   
        The validation process, illustrated in Figure \ref{fig:control_validation}, involves computing the average expected return of the current control policy and designs sampled from the current design distribution every iteration over the full validation dataset, i.e., $T=672$ hours. The validation performance of the co-optimisation scenario (blue) is benchmarked against the best two-step scenario (orange), the design-only scenario (green), and the fixed scenario (black). The experiments are conducted using 30 different seed values, with the median and quartiles reported in Figure \ref{fig:control_validation}. All scenarios quickly converge to a unique solution for this specific validation episode over $T=672$ hours. Interestingly, the difference in performance between the co-optimisation and best two-step scenarios is greater than during the training process. This might result from the perfect foresight approach in the MILP parameter selection, which allows for selection based on future information during learning that is unknown at the time of evaluation. As reported in Table \ref{tab:perf}, the scenarios converge to the following average values at the last iteration: $-50.1$ for co-optimisation, $-54.5$ for best two-step, $-97.1$ for design-only, and $-99.6$ for the fixed scenario.      
    
\begin{figure}[ht]
        \vskip 0.2in
        \begin{center}
        \centerline{\includegraphics[width=\columnwidth]{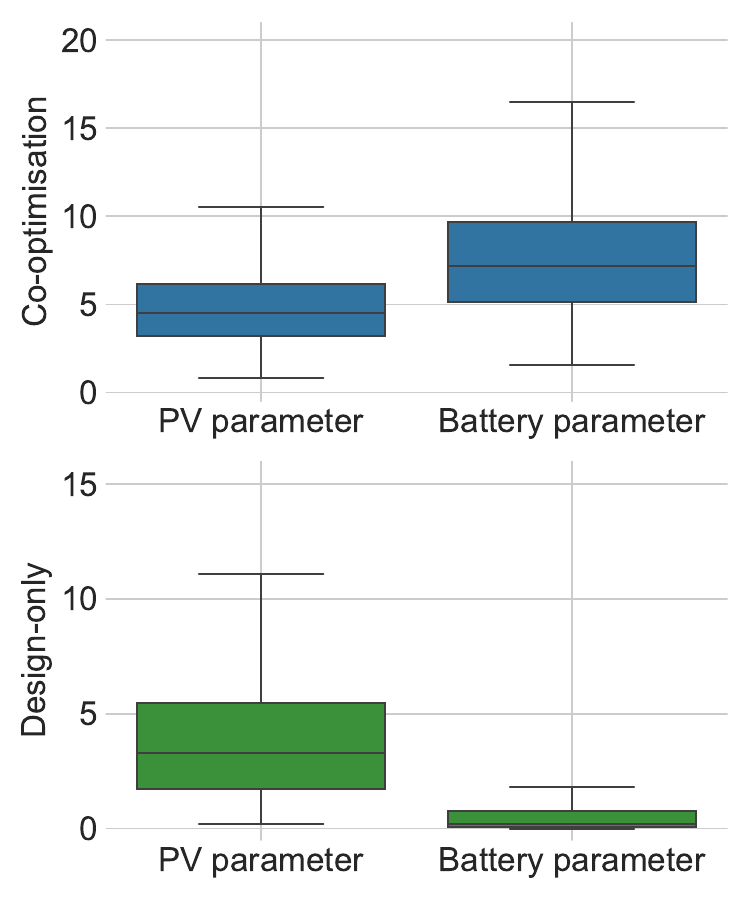}}
        \caption{Design parameter distribution after training for the co-optimisation (top, blue) and design-only (bottom, green) scenarios. The boxplots are computed based on a sample of 1000 designs drawn from the final design distribution of one of the 30 seed experiments.}
        \label{fig:boxplots}
        \end{center}
        \vskip -0.2in
        \end{figure}

        Figure \ref{fig:boxplots} represents the final distribution (estimated with 1000 samples) after 500 training iterations of one of the 30 seed experiments, for both the co-optimisation scenario and the design-only scenario. The median and quartiles are used to highlight the narrow confidence interval of the parameter distribution within the design space \(X = \mathbb{R}^+ \). This illustration effectively shows that both scenarios converge to similar optimal design parameter intervals. For co-optimisation with DDPG (blue), the interval between the first and third quartile is $\left[3, 6\right]$ kWp for PV and $\left[5, 10\right]$ kWh for battery capacity. For the design-only scenario with the rule-based control policy (green), the interval between the first and third quartile is $\left[2, 5\right]$ kWp for PV and $\left[0, 1\right]$ kWh for battery capacity. The mean design parameter over all 30 scenarios is reported in Table \ref{tab:design}. Note that these design parameter values are consistent with the assumptions of the building-scale system environment. Additionally, they differ from those computed using MILP, which in the best two-step scenario are equivalent to 6 kWp for PV and 14 kWh for battery capacity, reflecting different optimisation dynamics. 

        \begin{table}[ht]
            \caption{Mean of the design distribution at the last iteration, averaged over the 30 seed experiments in the co-optimisation and design-only scenarios. For the best two-step and fixed scenarios, the reported values are those computed using MILP.}
            \label{tab:design}
            \vskip 0.15in
            \begin{center}
            \begin{small}
            \begin{sc}
            \begin{tabular}{lcc}
            \toprule
            Design parameter & PV & Battery \\
            \midrule
            \textbf{Co-optim.}   & \textbf{6.6$\pm$1.4}  & \textbf{4.5$\pm$0.9}  \\
            Best 2-step & 6  & 14   \\
            Design-only &  6.3$\pm$5.3 & 3.5$\pm$12.2   \\
            Fixed       & 3  & 14  \\
            \bottomrule
            \end{tabular}
            \end{sc}
            \end{small}
            \end{center}
            \vskip -0.1in
        \end{table}

\section{Discussion}
\label{ch7:discussion}
    This study investigates the co-optimisation of design and operation in energy systems using a novel RL framework. The primary goal was to assess the feasibility and effectiveness of RL in developing integrated design strategies within a co-optimisation framework, aiming to enhance system performance by minimising total electricity costs. The results confirm that the framework successfully converges to high-performing design parameters while achieving superior control performance in both short and long-term periods. The co-optimisation scenario outperforms both the design-only and best two-step scenarios in training and validation performances, while converging to different design parameter values.
    
    First, the two RL-based design optimisations, i.e., the co-optimisation and design-only scenarios, converged to different design parameters while achieving significantly different operational performances, underscoring the significance of co-optimisation. The convergence to optimal design parameters in both scenarios is evidenced by the narrowing of the boxplot charts, indicating a non-dispersed solution. The optimised design parameters, although modest, align realistically with the environmental conditions and model assumptions. Additionally, they should be considered in relation to electricity requirements (i.e., 2.5 kWh on average per hour). These results highlight the framework's capacity to provide precise solutions. 

    The choice of DDPG among actor-critic algorithms was motivated by its off-policy nature and suitability for environments with continuous action spaces. These characteristics make DDPG significantly more sample-efficient, an advantage in the energy sector where system designs are typically based on data from a single year or, at best, a few years. The fast convergence of the control parameters further confirms the suitability of this algorithm for energy applications.

    The main limitation of this proposed framework is the difficulty in guaranteeing an optimal design, in contrast to the one computed using MILP. The results can vary due to sensitivity to hyperparameters, necessitating a detailed analysis of the evolution to ensure convergence to an optimal solution. This becomes even more complex when different algorithms are used and converge to different solutions, owing to the sensitivity to hyperparameters that must be carefully studied. Additionally, this study examines two design parameters. Scaling up the method to include additional design parameters presents two main challenges. First, it increases the difficulty in sampling interesting design spaces for all parameters, likely requiring more iterations. Second, it results in higher variance in the gradient estimates. This is analogous to problems where the optimal control policy is learned. 
    
    Finally, two important advantages from the energy perspective are: first, the framework provides an interval of optimal design values rather than a unique solution, as MILP typically does, offering more flexibility and sensitivity information. Second, this framework offers better performance without assuming perfect foresight, likely explaining the superior validation performance in Figure \ref{fig:control_validation}, as the MILP did not have access to the validation dataset while computing the design parameters.

\section{Conclusion}
\label{ch7:conclusion}
    The primary achievement of this study has been leveraging theoretical advances in RL to bridge the gap with practical energy challenges, focusing on the co-optimisation of design and operation within energy systems. This work has harnessed recent developments in policy gradient techniques to introduce an integrated, off-policy, and model-free RL framework tailored to tackle the co-optimisation challenge in energy systems.

    The successful demonstration of RL's feasibility and effectiveness in developing integrated design strategies within a co-optimisation framework paves the way for future research and expands the capabilities of RL in the energy sector. This conclusion aligns with two notable reviews: \cite{dranka_review_2021}, which underscores the importance of addressing co-optimisation in energy and highlights the absence of integrated solutions, and \cite{perera_applications_2021}, which notes that RL is not fully exploited in energy and suggests that using RL for design would be a promising new research area.

    The outcomes validate the relevance of using RL to design energy systems, demonstrating how co-optimisation can effectively compute control and design policies jointly, and surpass traditional approaches. Additionally, this framework does not mandate a specific control algorithm or restrict to RL alone, instead, it requires the problem to be formulated as an MDP. Adherence to RL standards, i.e., Gymnasium library \cite{towers_gymnasium_2023}, is advised to ensure seamless integration with existing control algorithms, even though they have been developed from scratch in this case.

    The practical application reveals the framework potential through a single year's data analysis. For greater accuracy and to evaluate long-term co-optimisation effects, it is advantageous to extend the dataset to encompass multiple years. Expanding the dataset would enhance the framework's ability to manage annual fluctuations in energy supply and demand. Further complexity could be introduced into the energy system model, like integrating multiple electric vehicles and accounting for non-linear heat pump dynamics. Additionally, incorporating more complex energy system dynamics such as real-time pricing or demand-response capabilities could improve the model's precision and relevance. The framework has also demonstrated promising outcomes that suggest the potential for generalisation to enhance sim-to-real transfer \cite{peng_sim--real_2018}, a significant step towards ensuring that the insights and predictions generated can be effectively applied in realistic operational settings \cite{schaff_sim--real_2023}. Future directions might also include integrating a critic architecture directly into the design learning process and extend the off-policy nature to the design part.

    In conclusion, the findings and the comparison to traditional approaches, such as the design-only and best two-step scenarios, highlight that optimal design and optimal control are intrinsically linked. These insights affirm the value of integrated co-optimisation strategies over traditional, segregated approaches, especially in complex and dynamic settings like modern energy systems.

\section*{Acknowledgements}
The authors would like to thank Prof. Gilles Louppe for providing access to the Alan clusters, which facilitated the experiments in this work. Adrien Bolland gratefully acknowledges the financial support of a research fellowship of the F.R.S.-FNRS.

\section*{Impact Statement}
This paper presents work whose goal is to advance the field of Machine Learning. There are many potential societal consequences of our work, none which we feel must be specifically highlighted here.


\bibliography{main}
\bibliographystyle{icml2024}

\newpage
\appendix
\onecolumn
\section{Appendix Building-Scale System -- Environment definition}
\label{ap:A}

        This Annex details the building-scale energy system used within an office setting, equipped with a PV  (Photovoltaic) installation and a stationary lithium-ion battery to satisfy its electricity requirements. The system also features a bidirectional EV (Electric Vehicle) charging point, whose usage is stochastically modelled based on typical patterns. Moreover, the building is connected to the electrical grid, subject to dynamically varying electricity prices. The main objective is to determine the optimal design for the PV installation (\(P^\textsc{nom}\)) and the battery capacity (\(B\)), while simultaneously developing an optimal control policy for battery and EV management. This aims to minimise the total cost of ownership, encompassing both capital and operational expenses, as well as grid costs. The environment is formulated below as an MDP and Table \ref{tab_ch7:constants} gathers all parameters of this environment.

        \begin{table}[h!]
        	\centering
        	\small
        		\begin{tabular}{@{}llllll@{}}
        			\toprule
        			& \textbf{Parameter} & \textbf{Value} & \textbf{Set} & \textbf{Unit}  & \textbf{Description}  \\
        			\midrule
        			\parbox[t]{2mm}{\multirow{5}{*}{\rotatebox[origin=c]{90}{\textsc{Grid}}}}
                            &$P^\textsc{imp}$ & & $\mathbb{R}_+^T$ & kW & imported power (from the grid)\\
		                  &$P^\textsc{exp}$ & & $\mathbb{R}_+^T$ & kW & exported power (to the grid)\\
            			&$C^\textsc{imp}_\textsc{grid}$ &  & $\mathbb{R}^T$ & CHF/kWh & imported electricity price \\ 
            			&$C^\textsc{exp}_\textsc{grid}$ &  & $\mathbb{R}^T$ & CHF/kWh & exported electricity price \\
            		    &$C_\textsc{grid}$ &  & $\mathbb{R}^T$ & CHF & total electricity grid cost \\
        			\midrule
        			\parbox[t]{2mm}{\multirow{11}{*}{\rotatebox[origin=c]{90}{\textsc{PV}}}}
                            &$P^\textsc{nom}$ & & $\mathbb{R}_+$ & kW$_p$ & nominal power of the PV installation \\
            			&$P^\textsc{nom}_\textsc{min}$ & 0 & $\mathbb{R}_+$ & kW$_p$  & minimal nominal PV power \\ 
            			&$P^\textsc{nom}_\textsc{max}$ & $\infty$ & $\mathbb{R}_+$ & kW$_p$  & maximal nominal PV power \\			
            			&$P^\textsc{prod}$ &  & $\mathbb{R}_+^T$ & kW & generated PV power \\
                            &$p^\textsc{prod}$ &  & $\mathbb{R}_+^T$ & kW & normalised PV power \\
                    	  &$L^\textsc{pv}$  & 20 & $\mathbb{N}$ & years & PV lifetime\\
                            &$\textsc{R}_\textsc{pv}$ &  & $\mathbb{R}_+$ & - & annuity factor\\ 
            			&$\textsc{ox}_\textsc{pv}^\textsc{fix}$  & 0 & $\mathbb{R}_+$ & CHF & \textsc{opex} PV fixed cost\\  
            			&$\textsc{ox}_\textsc{pv}^\textsc{var}$ & 100 & $\mathbb{R}_+$ & CHF/kW & \textsc{opex} PV variable cost\\
            			&$\textsc{cx}_\textsc{pv}^\textsc{fix}$ & 100 & $\mathbb{R}_+$ & CHF & \textsc{capex} PV fixed cost\\
            			&$\textsc{cx}_\textsc{pv}^\textsc{var}$ & 775 & $\mathbb{R}_+$ & CHF/kW & \textsc{capex} PV variable cost\\ 
        			\midrule
        			\parbox[t]{2mm}{\multirow{12}{*}{\rotatebox[origin=c]{90}{\textsc{Battery}}}}
                            &$B$ & & $\mathbb{R}_+$ & kWh & nominal capacity of the battery \\
                		&$\textsc{soc}$ & & $\mathbb{R}_+^T$ & kWh & state of charge of the battery \\
                		&$P^B$ & & $\mathbb{R}^T$ & kW & power exchanged with the battery \\
                            &$B_\textsc{min}$ & 0 & $\mathbb{R}_+$ & kWh  & minimal nominal battery capacity \\ 
            			&$B_\textsc{max}$ & $\infty$ & $\mathbb{R}_+$ & kWh  & maximal nominal battery capacity \\
            			&$\eta^\textsc{b}$ & 0.9 & $\left]0,1\right]$ & - & battery  efficiency\\
                            &$L^\textsc{b}$ & 10 & $\mathbb{N}$ & years & battery lifetime\\
                            &$\textsc{R}_\textsc{B}$ &  & $\mathbb{R}_+$ & - & annuity factor\\
                            &$\textsc{ox}_\textsc{b}^\textsc{fix}$ & 0 & $\mathbb{R}_+$ & CHF & \textsc{opex} Battery fixed cost\\  
            			&$\textsc{ox}_\textsc{b}^\textsc{var}$ & 10 & $\mathbb{R}_+$ & CHF/kW & \textsc{opex} Battery variable cost\\
            			&$\textsc{cx}_\textsc{b}^\textsc{fix}$ & 50 & $\mathbb{R}_+$ & CHF & \textsc{capex} Battery fixed cost\\
            			&$\textsc{cx}_\textsc{b}^\textsc{var}$ & 300 & $\mathbb{R}_+$ & CHF/kW & \textsc{capex} Battery variable cost\\ 
        			\midrule
                        \parbox[t]{2mm}{\multirow{9}{*}{\rotatebox[origin=c]{90}{\textsc{EV}}}}
                            &$b^\textsc{ev}$ & & $\mathbb{R}_+^T$ & -  & binary indicator of EV presence \\
                            &$B^\textsc{ev}$ & 80 & $\mathbb{R}_+$ & kWh  & maximal nominal EV battery capacity \\
                            &$\textsc{soc}^\textsc{ev}$ & & $\mathbb{R}_+^T$ & kWh & state of charge of the EV battery \\
                            &$\textsc{soc}^\textsc{ev}_\textsc{min}$ & 32 & $\mathbb{R}_+$ & kWh & minimum state of charge of the EV battery \\
                            &$P^\textsc{ev}$ &  & $\mathbb{R}_+^T$ & kW & power exchange with the EV battery \\
                            &$P^\textsc{ev}_\textsc{max}$ & 5 & $\mathbb{R}_+$ & kW & maximal power exchange with the EV battery \\                            
            			&$\eta^\textsc{ev}$ & 1 & $\left]0,1\right]$ & - & EV battery efficiency\\
                            &$C^\textsc{imp}_\textsc{ev}$ & -1.5 & $\mathbb{R}$ & CHF/kWh & imported electricity price from the EV battery \\ 
            			&$C^\textsc{exp}_\textsc{ev}$ & 1 & $\mathbb{R}$ & CHF/kWh & exported electricity price to the EV battery \\ 
        			\midrule
        			\parbox[t]{2mm}{\multirow{4}{*}{\rotatebox[origin=c]{90}{\textsc{System}}}} 
            			&$T$  &  & $\mathbb{N}$& -& time horizon \\ 
            			&$\Delta t$ & 1 &  $\mathbb{R}_+$ & h  & time steps \\
            			& $r$ & 0.05 & $\mathbb{R}$ & -  & discount rate \\ 
            			& $P^\textsc{load}$ &  &  $\mathbb{R}_+^T$ & kW & uncontrollable electricity consumption\\
        			\bottomrule
        		\end{tabular}
            \caption{Set of constants and parameters of the building-scale PV-battery system studied.}
        	\label{tab_ch7:constants}
        \end{table}

        The \textbf{State Space} of the system can be fully described by
        \begin{align}
            s_t &= (h_t, d_t, \textsc{soc}_t,  P^\textsc{prod}_t,  P^\textsc{load}_t, C^\textsc{imp}_{\textsc{grid},t}, C^\textsc{exp}_{\textsc{grid},t}, b^\textsc{ev}_t,  \textsc{soc}^\textsc{ev}_t) \in \mathcal{S}
        \end{align}
        \begin{itemize}
            \item $h_t \in \{0, ..., 23\}$ denotes the hour of the day at time $t$.
            \item $d_t \in \{0, ..., 364\}$ denotes the day of the year at time $t$. 
            \item $\textsc{soc}_t \in [0, B]$ is the state of charge of the battery at time $t$, this value is upper bounded by the nominal capacity of the installed battery $\textsc{B}$. 
            \item $P^\textsc{prod}_t \in \mathbb{R}_+$ represents the expected PV power at time $t$. This value is obtained by scaling normalized historical data $p^\textsc{prod}_t$ with the design of PV power ($P^\textsc{nom}$) and considering $h_t$ and $d_t$ values.
            \item $P^\textsc{load}_t \in \mathbb{R}_+$ denotes the expected value of the electrical load at time $t$. The load profile is determined using historical data that corresponds to the same hour and day as the PV power.
            \item $C^\textsc{imp}_{\textsc{grid},t} \in \mathbb{R}$ represents the cost per unit of electricity imported from the grid at time $t$. This value is dynamically determined from a predefined dataset. 
            \item $C^\textsc{exp}_{\textsc{grid},t} \in \mathbb{R}$ corresponds to the compensation received per unit of electricity exported to the grid at time $t$. Like the import costs, this value is derived from a dataset. 
            \item $b^\textsc{ev}_t \in \{0,1\}$ is a binary indicator indicating whether a bidirectional EV is present at the charging station at time $t$. This state affects the potential for energy storage or retrieval from the EV's battery, thereby influencing the overall energy management strategy. The value is updated according to usage patterns captured in the dataset.
            \item $\textsc{soc}^\textsc{ev}_t \in [\textsc{soc}^\textsc{ev}_\textsc{min},B^\textsc{ev}]$ specifies the current charge level of the EV's battery, when present. This value ranges between 40\,\% of $B^\textsc{ev}$ and $B^\textsc{ev}$ when the EV is connected, and is set to zero when no EV is present. The charge level is initialised randomly based on probable starting conditions and adjusted according to actual charging and discharging activities dictated by the control policy and EV usage scenarios from the dataset.
        \end{itemize}

        The \textbf{Action Space} comprises the power exchanged with the stationary battery and the EV's battery when present. Positive values indicate discharging, and negative values represent charging. The continuous action space is defined as:
        \begin{align}
            a_t &= (\widetilde{P}_t^B, \widetilde{P}_t^\textsc{ev}) \in \mathcal{A} = [-\frac{B}{\Delta t}, \frac{B}{\Delta t}] \times [-P^\textsc{ev}_\textsc{max}, P^\textsc{ev}_\textsc{max}]
        \end{align}

        The \textbf{Initial Distribution} set the initial state as follows. The hour $h_0$ is set to 0. During the training process, the initial day $d_0$ is randomly selected, whereas for the validation process, $d_0$ is set to the earliest date within the year. The initial $\textsc{soc}_t$ is randomly determined during training and set to half of the battery capacity $B$ during validation. All other initial state values are derived from an predefined input dataset based on the corresponding initial hour and day.

        The \textbf{Transition Probability} becomes a transition function, as there is no randomness involved. This function updates the system state at each hourly time step.
        
        The hour of the day $h_t$ increments each hour, and the day $d_t$ increments every 24 hours:
        \begin{align}
            \label{eq:h+1}
            h_{t+1} &= (h_t + 1) \text{ mod } 24\\
            d_{t+1} &= \text{Int}(\frac{h_t + 1}{24})
            \label{eq:d+1}
        \end{align}
        \noindent where the function $Int$ takes the integer value of the expression.

        The state of charge for both the stationary battery $\textsc{soc}_t$ and the EV's battery $\textsc{soc}^\textsc{ev}_t$ are updated based on the respective power actions $\widetilde{P}_t^B$ and $\widetilde{P}_t^\textsc{ev}$. These actions specify the power to be charged or discharged from the batteries over one hour ($\Delta t = 1h$). However, the actual power exchanged is constrained either by the battery capacity when charging it or by the energy stored in the battery when discharging it.
        \begin{align}
            P^B_t = 
            \begin{cases}
              \frac{\textsc{B} - \textsc{soc}_t}{\Delta t} &\text{ if } \widetilde{P}^B_t > \frac{\textsc{B} - \textsc{soc}_t}{\Delta t}\\
              \frac{\textsc{soc}_t}{\Delta t} &\text{ if } \widetilde{P}^B_t < - \frac{\textsc{soc}_t}{\Delta t} \\
              P^B_t &\text{otherwise}
            \end{cases}
        \end{align}
        Similarly for the EV's battery:
        \begin{align}
            P^\textsc{ev}_t &= 
                \begin{cases}
                  \frac{B^\textsc{ev} - \textsc{soc}^\textsc{ev}_t}{\Delta t} &\text{ if } \widetilde{P}^\textsc{ev}_t > \frac{B^\textsc{ev} - \textsc{soc}_t^\textsc{ev}}{\Delta t}\\
                  \frac{\textsc{soc}^\textsc{ev}_t}{\Delta t} &\text{ if } \widetilde{P}^\textsc{ev}_t < - \frac{40\% \cdot B^\textsc{ev}_t}{\Delta t} \\
                  P^\textsc{ev}_t &\text{otherwise}
                \end{cases}\\
            &\text{with} 
            -P^\textsc{ev}_\textsc{max} \leq \widetilde{P}_t^\textsc{ev} \leq P^\textsc{ev}_\textsc{max}
        \end{align}
        Using these power exchanges, the state of charge for the next time step is calculated as:
        \begin{align}
            \textsc{soc}_{t+1} &= \textsc{soc}_t + P^B_t \cdot \Delta t \cdot (\eta^\textsc{b} \text{ if } P^B_t \geq 0 \text{ else } \frac{1}{\eta^\textsc{b}})\\
            \textsc{soc}^{\textsc{ev}}_{t+1} &= \textsc{soc}^{\textsc{ev}}_t + P^{\textsc{ev}}_t \cdot \Delta t \cdot (\eta^\textsc{ev} \text{ if } P^{\textsc{ev}}_t \geq 0 \text{ else } \frac{1}{\eta^\textsc{ev}})
        \end{align}
        where $\eta^\textsc{b}$ and $\eta^\textsc{ev}$ are respectively the efficiency of the battery and EV's battery. 
        

        The \textbf{Reward Function} quantifies the system's performance by incorporating economic factors that include investment cost (\textsc{capex}), operating cost (\textsc{opex}), and costs associated with the purchase and sale of electricity from the grid. The reward at each time step $t$ is calculated as the negative total expenditure (\textsc{totex}):
        \begin{align}
            r_t &= -\textsc{totex}_t \\
            &= - (\textsc{capex} + \textsc{opex} + C_{\textsc{grid},t}) \\
            &= - (\textsc{capex} + \textsc{opex} + P^\textsc{imp}_t \cdot C^\textsc{imp}_{\textsc{grid},t} - P^\textsc{exp}_t \cdot C^\textsc{exp}_{\textsc{grid},t})
        \end{align}
        where $C_{\textsc{grid},t}$ represents the net cost of electricity exchanged with the grid at time $t$.
        
        The total cost (\textsc{totex}) includes:
        \begin{align}
            \textsc{totex} &= \textsc{opex} + \textsc{capex} + C_\textsc{grid}
        \end{align}
        Operating costs (\textsc{opex}) and capital expenditure (\textsc{capex}) are defined for both PV and battery design parameters as:
        \begin{align}
            \textsc{opex} &= \textsc{ox}_\textsc{pv} + \textsc{ox}_\textsc{B} \\
            \textsc{capex} &= \textsc{cx}_\textsc{pv} \cdot R_{pv} + \textsc{cx}_\textsc{B} \cdot R_{B}
        \end{align}
        where $R_\textsc{pv}$ and $R_{B}$ are annuity factors adjusting the \textsc{capex} for the lifetime of the system's components, considering their financial amortisation over a finite period $T$.
        
        The annuity factor $R$ is derived as follows to prorate the \textsc{capex} over the operational duration $T$, acknowledging $T$ in hours and 8760 as the number of hours in a year:
        \begin{align}
            R = \frac{r \cdot (1 + r)^L}{(1 + r)^L - 1} \cdot \frac{T}{8760}
        \end{align}
        This factor is calculated using the annual discount rate $r$ and the expected lifetime $L$ of the components, thereby aligning the investment costs proportionally to the duration $T$ of the optimisation horizon. 

        \begin{table}[h!]
            \centering
            \small
            \caption{Synthetic dataset of electricity pricing and EV arrival time of the building-scale PV-battery system studied.}
            \begin{tabular}{@{}ccccc@{}}
                \toprule
                & \textbf{h} & \textbf{$C^\textsc{exp}_\textsc{grid}$} & \textbf{$C^\textsc{imp}_\textsc{grid}$} & \textbf{Probability of EV arrival time}  \\
                & [-] & [CHF/kWh] & [CHF/kWh] & [-]  \\
                \midrule
                & 0 & 0 & -0.3 & 0 \\
                & 1 & 0 & -0.3 & 0 \\
                & 2 & 0 & -0.3 & 0 \\
                & 3 & 0 & -0.3 & 0 \\
                & 4 & 0 & -0.3 & 0 \\
                & 5 & 0 & -0.3 & 0 \\
                & 6 & 0 & -0.5 & 0 \\
                & 7 & 0 & -0.5 & 0.75 \\
                & 8 & 0 & -0.5 & 0.9 \\
                & 9 & 0 & -0.5 & 0.9 \\
                & 10 & 0 & -0.3 & 0.75 \\
                & 11 & 0 & -0.3 & 0.1 \\
                & 12 & 0 & -0.3 & 0.1 \\
                & 13 & 0 & -0.3 & 0.1 \\
                & 14 & 0 & -0.3 & 0 \\
                & 15 & 0 & -0.3 & 0 \\
                & 16 & 0 & -0.5 & 0 \\
                & 17 & 0 & -0.5 & 0 \\
                & 18 & 0 & -0.5 & 0 \\
                & 19 & 0 & -0.5 & 0 \\
                & 20 & 0 & -0.5 & 0 \\
                & 21 & 0 & -0.5 & 0 \\
                & 22 & 0 & -0.3 & 0 \\
                & 23 & 0 & -0.3 & 0 \\
                \bottomrule
            \end{tabular}
            \label{tab_ch7:data}
        \end{table}
        
        \begin{figure}[ht]
            \centering
            \includegraphics[width=1\linewidth]{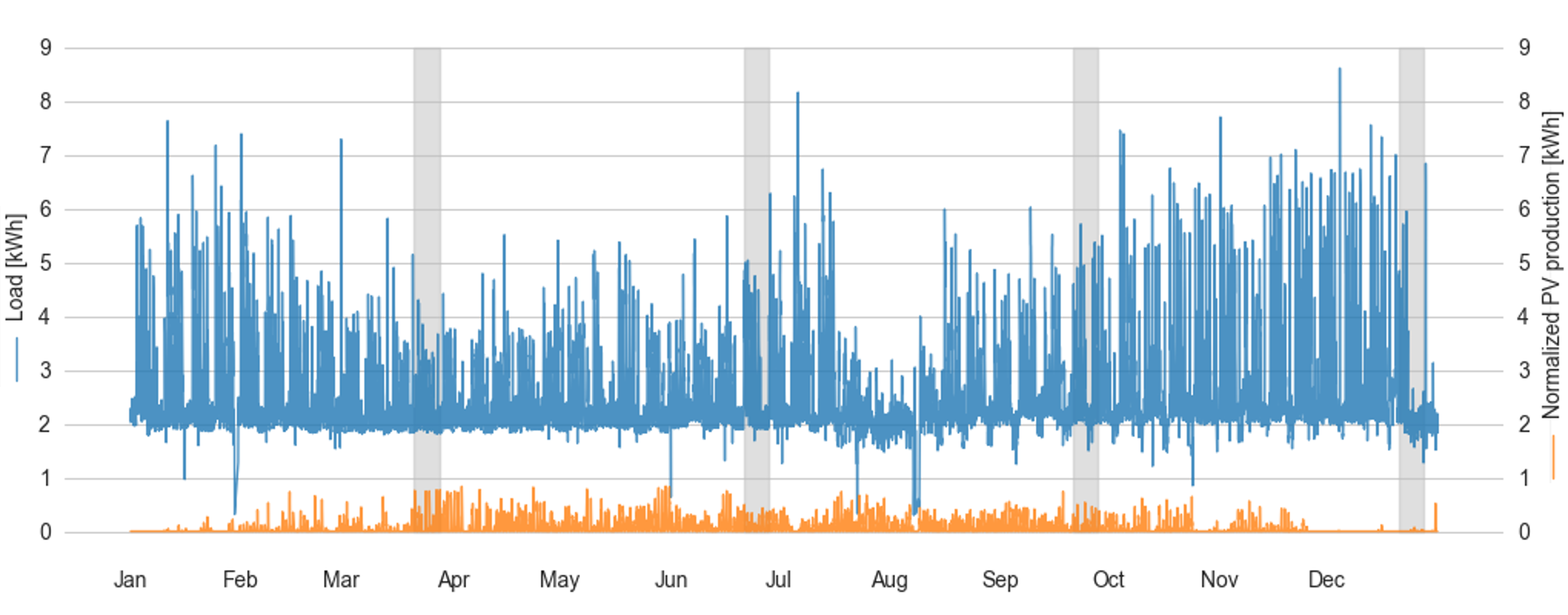}
            \caption{Visualisation of the historical dataset covering a year of the building electricity consumption and its normalised PV production. The white background indicates the training set, while the grey background represents the validation dataset.}
            \label{fig:load_pv}
        \end{figure}

        The \textbf{Optimisation Horizon} refers to the period over which the system is optimised, corresponding to the duration of an episode. In this model, each time step of the MDP represents a single hour, and the horizon is truncated after $T=168$ hours, equivalent to one week. Long-term dependencies are captured through the bootstrapping method used to train the critic. However, the time horizon ideally would span an entire year, or even the full lifecycle of the system to capture seasonal fluctuations in production and consumption, as well as potential equipment degradation. To assess performance over such a longer time horizon, the performances are regularly evaluated during the training phase across the full training dataset, corresponding to $T=8088$. 

        The \textbf{Historical Datasets} of the system are detailed in Table \ref{tab_ch7:constants}. The historical data for the normalised PV production and the electrical consumption are derived from real monitoring of an office building in Switzerland in 2021, as shown in Figure \ref{fig:load_pv}. This dataset is divided into training and validation parts, each selected to represent the seasonal fluctuations. The dataset used for the electricity prices supplied to and from the grid, as well as the arrival times of the EV, are synthetically generated and summarised in Table \ref{tab_ch7:data}. The grid export cost, $C^\textsc{exp}_\textsc{grid}$, is set to 0 at all times to discourage making money by reselling PV production and to maximise self-consumption. The duration of the EV's presence is randomly varied between 5 and 8 hours, and the initial state of charge (SoC) of the EV is randomly set between 40\,\% and 100\,\% of its battery capacity, $B^\textsc{ev}$.


\end{document}